\documentclass[runningheads]{llncs}
\usepackage[usenames,dvipsnames]{xcolor}
\usepackage{graphicx}
\usepackage{hyperref}
\usepackage[normalem]{ulem}
\useunder{\uline}{\ul}{}
\usepackage[font={footnotesize}]{caption}
\begin{document}
\title{Region-of-interest guided Supervoxel Inpainting for Self-supervision}
\titlerunning{Region-of-interest guided Supervoxel Inpainting for Self-supervision}
%
%
%
%
%

\author{
Subhradeep Kayal\inst{1}, Shuai Chen\inst{1} \and Marleen de Bruijne\inst{1,2}
}

\institute{
Erasmus MC, Biomedical Imaging Group Rotterdam, Departments of Radiology and Nuclear Medicine, The Netherlands \and
University of Copenhagen, Machine Learning Section, Department of Computer Science, Denmark
}

\authorrunning{S. Kayal et al.}

\maketitle              
\begin{abstract}
Self-supervised learning has proven to be invaluable in making best use of all of the available data in biomedical image segmentation. One particularly simple and effective mechanism to achieve self-supervision is inpainting, the task of predicting arbitrary missing areas based on the rest of an image. In this work, we focus on image inpainting as the self-supervised proxy task, and propose two novel structural changes to further enhance the performance of a deep neural network. We guide the process of generating images to inpaint by using supervoxel-based masking instead of random masking, and also by focusing on the area to be segmented in the primary task, which we term as the region-of-interest. We postulate that these additions force the network to learn semantics that are more attuned to the primary task, and test our hypotheses on two applications: brain tumour and white matter hyperintensities segmentation. We empirically show that our proposed approach consistently outperforms both supervised CNNs, without any self-supervision, and conventional inpainting-based self-supervision methods on both large and small training set sizes.
\end{abstract}
\section{Introduction and Motivation}



\emph{Self-supervised learning} points to methods in which neural networks are explicitly trained on large volumes of data, whose labels can be determined automatically and inexpensively, to reduce the need for manually labeled data. Many ways of performing self-supervision exist, amongst which a popular way is the \emph{pre-train and fine-tune} paradigm where: (1) a convolutional neural network is pre-trained on a proxy task for which labels can be generated easily, and (2) it is then fine-tuned on the main task using labeled data. Utilizing a suitable and complex proxy task, self-supervision teaches the network robust and transferable visual features, which alleviates overfitting problems and aides its performance when fine-tuned on the main task \cite{DBLP:journals/corr/abs-1902-06162}.

In the medical imaging domain a variety of proxy tasks have been proposed, such as sorting 2D slices derived from 3D volumetric scans \cite{7950587}, predicting 3D distance between patches sampled from an organ \cite{DBLP:conf/miccai/SpitzerKAHD18}, masking patches or volumes within the image and learning to predict them \cite{CHEN2019101539}, and shuffling 3D blocks within an image and letting a network predict their original positions \cite{DBLP:conf/miccai/ZhuangLHMYZ19}. Recently, state-of-the-art results were achieved on several biomedical benchmark datasets by networks which were self-supervised using a sequence of individual proxy tasks \cite{10.1007/978-3-030-32251-9_42}.


\begin{figure}[!t]
  \centering
    \includegraphics[width=1.0\textwidth]{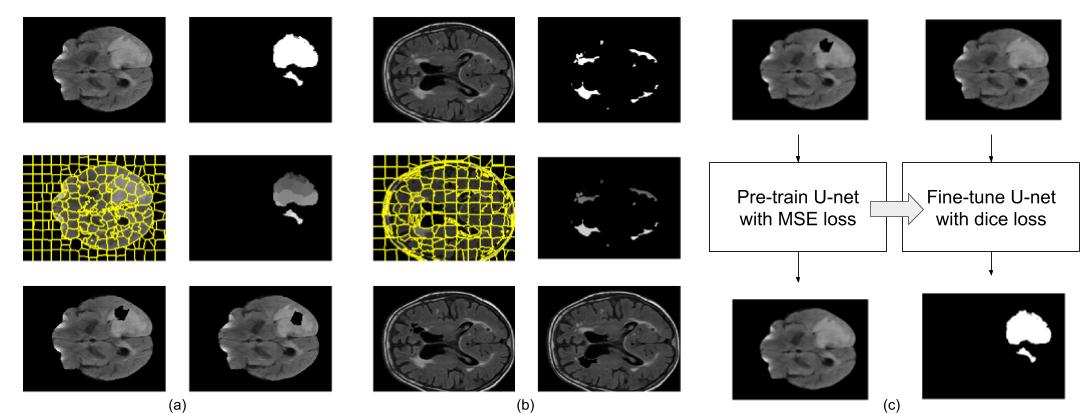}
    \caption{\textbf{Proposed ROI-guided inpainting.} (a) Examples from BraTS 2018 dataset (left to right from top to bottom): original FLAIR image-slice, ground-truth segmentation map, FLAIR image-slice with superpixels overlaid, region-of-interest (ROI) influenced superpixels, examples of synthesized images to be inpainted. (b) Examples from White Matter Hyperintensities 2017 dataset. Notice that the ground-truth segmentations are much smaller in size. (c) first a U-net is pre-trained on the inpainting task with MSE loss, next it is fine-tuned on the main segmentation task with Dice loss.}
    \label{figinpainting}
\end{figure}

Prior works in self-supervision literature have designed the proxy task largely uninfluenced by the downstream task in focus. However, since the features that the network learns are dependent on where it is focusing on during the self-supervision task, it might be beneficial to bias or \emph{guide} the proxy task towards areas that are of interest to the main task. Specifically for image segmentation, these would be the foreground areas to be segmented in the main task, which we term as the \emph{region-of-interests or ROIs}.

We experiment with the proxy task of inpainting \cite{DBLP:conf/cvpr/PathakKDDE16}, where the network must learn to fill-in artificially created gaps in images. In the context of biomedical imaging, a network that learns to inpaint healthy tissue will learn a different set of semantics than one which inpaints various kinds of tumours. Thus, if the main task is that of segmenting tumours, it can be hypothesized that having a network inpaint tumourous areas as a proxy task will likely teach it semantics attuned to segmenting tumours, and thereby be more beneficial for the main task than learning general semantics. In other words, by increasing the frequency of inpainting tumours, we can teach the network features which are more related to the tumour segmentation task.

Furthermore, in prior work the selection of regions to mask has largely been uninformed and random. We try to improve upon this situation by selecting regions which are homogeneous. Masking such regions could force the network to learn more about the anatomical meaning and relation to other structures of the masked tissue. For example, masking small regions in a lung CT scan would only require the model to correctly interpolate the structures (airways, vessels) around the masked region. In contrast, when a full airway or vessel branch is masked, inpainting requires understanding of the relation between branches in vessel or airway trees and/or the relation between airways and arteries, a piece of information that has been found to improve airway segmentation \cite{LO2010527}.

The contributions of this work are stated next. Firstly, this paper demonstrates that guiding the inpainting process with the main class(es) of interest (,i.e., the segmentation foreground, interchangeably used with the term \emph{ROI} in this paper) during the self-supervised pre-training of a network improves its performance over using random regions. Secondly, we show that instead of inpainting regions of regular shapes in an uninformed way, further performance gain is possible if the masked regions are chosen to be homogeneous. This is done by constructing supervoxels, in a preprocessing step, and using them as candidate regions to be inpainted. In order to show the efficiency of these proposed changes, we conduct empirical analyses on two popularly used public datasets for biomedical image segmentation.

\section{Methods}
\label{methodology}

The proposed method (Figure \ref{figinpainting}) utilizes supervoxelization to create candidate regions, followed by selecting only those supervoxels which have an overlap with (any of) the foreground class(es). The selected supervoxels are utilized in the inpainting process, where we use them as masks to suppress areas in an image to train a network to predict (or \emph{inpaint}) them based on their surroundings. Since we control the parameters of this process, it can be used to create an arbitrarily large amount of synthetic training images for pre-training.

\subsection{Region-of-interest guided inpainting}

Inpainting is an effective proxy task for self-supervision, which proceeds by training a network to reconstruct an image from a masked version of it. In this section, we explain our proposed masking approach, followed by the description of the network in Section \ref{trainingstrat}.

\noindent\textbf{Supervoxelization:} While previous works primarily use random grids and cubes as candidate regions to inpaint, the first step in our proposed approach is to select regions based on some notion of homogeneity. One way of achieving this is to construct supervoxels, which may be defined as homogeneous groups of voxels that share some common characteristics. A particularly efficient algorithm to construct such supervoxels is \emph{SLIC} or \emph{simple linear iterative clustering} \cite{6205760}.

For 3D medical images, SLIC can cluster voxels based on their intensity values, corresponding to the various modalities, and spatial coordinates of the voxel within the image. SLIC has two main hyperparameters: one, \emph{compactness}, controls the balance between emphasis on intensity values and spatial coordinates (larger values make square/cubic grids), and the other defines the maximum number of supervoxels. Examples in the second row of Figure \ref{figinpainting}, subfigure (a) and (b).

In this work, we use SLIC with intensity values corresponding to the two modalities we used in our experiments, FLAIR and T1 (or contrast enhanced T1), in order to construct supervoxels. The exact parameter settings for supervoxelization are described later in Section \ref{experiments:settings}.

\noindent\textbf{ROI-guided masking for inpainting image synthesis:} Once the supervoxel labels have been created, the next step is to retain only those supervoxels which have an overlap with the region-of-interest. To achieve this, we first convert the segmentation map to a binary one by considering all foreground areas to be a class with a label value as $1$ and the background as $0$, since there may be multiple regions-of-interest in a multi-class segmentation setting. Then an elementwise \emph{and} operation is performed between the resulting binary segmentation map and the generated supervoxel. For all the supervoxels that remain, training images for the inpainting task can be synthesized by masking an area corresponding to such a \emph{ROI-guided supervoxel}, with the original unmasked image being the target for the network. Some examples of this are in the second row of Figure \ref{figinpainting}, subfigure (a) and (b).

By constructing a training set for the inpainting task in this fashion, we are essentially increasing the frequency of inpainting regions which are important to the main task more than random chance. This is what, we posit, will bring about improvements in the performance of the network on the main task.

Formally, let $D_{train} = \{(I_i, S_i)\}_{i=1..n}$ be the training dataset containing $n$ images with $I_i$ being a 3D multi-modal training image and $S_i$ being the segmentation ground-truth label, containing zero values representing background. If $f$ is a supervoxelization algorithm (in our case, SLIC), then a ROI-guided supervoxelized image is given by ${R_i} = f(I_i) \odot S_i$, where $\odot$ signifies elementwise multiplication.

${R_i}$ contains supervoxel regions having non-zero labels corresponding to foregound supervoxels. Then, the synthetic dataset for inpainting, $D_{inp}$ is constructed as:
\begin{equation}
    D_{inp} = \Bigg\{ \Big\{ \big( I_i \odot r_{ij}^0, I_i \big) \Big\}_{r_{ij} \in R_i, j=1..{m_i}} \Bigg\}_{i=1..n}
\end{equation}
where $r_{ij}$ is a single supervoxel region in the set ${R_i}$, which contains a total of $m_i$ supervoxels, and $r_{ij}^0$ is the corresponding inverted region-mask, containing 0 for voxels belonging to the region and 1 everywhere else. $I_i \odot r_{ij}$ is then the masked image input to the network and $I_i$ is the expected output to reconstruct, the target for the inpainting task. Thus, the maximum cardinality of $D_{inp}$ can be $n \times m_i$.

Examples $ D_{inp}$ are in the last row of Figure \ref{figinpainting}, subfigure (a) and (b).

\subsection{Training Strategy}
\label{trainingstrat}

\noindent\textbf{Network:} For all the experiments, a shallow 3D U-net \cite{10.1007/978-3-319-24574-4_28,10.1007/978-3-319-46723-8_49} containing 3 resolution levels has been used, with a batch-normalization layer after every convolution layer. In our experiments we find that 3 layers provide sufficient capacity for both the inpainting and the segmentation task. Since we use two modalities for our experiments, the U-net has two input channels.

If we were to use an image reconstruction proxy task, a U-net would learn to copy over the original image because of its skip connections, and would not be useful in learning features. In our task of inpainting the network never sees the masked regions and, therefore, cannot memorize it, making the use of a U-net reasonable.

\noindent\textbf{Pre-training:} In order to pre-train the network, it is fitted to the $D_{inp}$ dataset by minimizing the mean squared error (MSE) between the masked and the original images using the \emph{Adam} \cite{DBLP:journals/corr/KingmaB14} optimizer. We call this model \emph{inpainter U-net}.

\noindent\textbf{Fine-tuning:} The inpainter U-net is then fine-tuned on the (main) segmentation task using the original labeled training dataset, $D_{train}$, by optimizing the Dice segmentation overlap objective on the labeled images. If the data is multi-modal, the inpainter U-net will be trained to produce multi-channel outputs, in which case we would need to replace the last 3D convolutional layer to have a single-channel output for segmentation.

More details about the network parameters are provided in section \ref{experiments:settings}.

\section{Experimental Settings}
\label{experiments}

\subsection{Data}
\label{experiments:data}
For our experiments, we use two public datasets containing 3D MRI scans and corresponding manual segmentations.

\noindent\textbf{BraTS 2018} \cite{8669968}: 210 MRI scans from patients with high-grade glioma are randomly split three times into 150, 30 and 30 scans for training, validation and testing, respectively, using a 3-fold Monte-carlo cross-validation scheme. To be able to easily compare our method against baselines, we focus on segmenting the whole tumour and use two of the four modalities, \emph{FLAIR} and \emph{T1-gd}, which have been found to be the most effective at this task \cite{10.1007/978-3-319-24553-9_65}.

\noindent\textbf{White Matter Hyperintensities (WMH) 2017} \cite{6975210}: The total size of the dataset is 60 FLAIR and T1 scans, coming from 3 different sites, with corresponding manual segmentations of white matter hyperintensities. We employ a 3-fold Monte-carlo cross-validation scheme again, splitting the dataset into 40, 10 and 10 for training, validation and testing, respectively, and use both of the available modalities for our experiments.

\subsection{Baseline Methods}
\label{experiments:baseline}

We term the technique proposed in this paper as \emph{roi-supervoxel} to denote the use of the segmentation map and supervoxelization to guide the inpainting process used for pre-training. In order to validate its effectiveness, it is tested against the following baselines: \emph{vanilla-unet}: a U-net without any pre-training; \emph{restart-unet}: a U-net pre-trained on the main (segmentation) task and fine-tuned on the same task for an additional set of epochs; \emph{noroi-grid}: the more traditional inpainting mechanism where random regular sized cuboids are masked; \emph{roi-grid}: a similar process as \emph{roi-supervoxel}, except for the use of regular cuboids overlapping with the segmentation map, instead of supervoxel regions, for masking; \emph{noroi-supervoxel}: where random supervoxels are masked.

\subsection{Settings}
\label{experiments:settings}

\noindent\textbf{Inpainting Parameters}: The inpainting process starts by creating the supervoxel regions using SLIC\footnote{We use the implementation in \url{https://scikit-image.org/docs/dev/api/skimage.segmentation.html?highlight=slic\#skimage.segmentation.slic}}. We fix these the compactness value at 0.15 and choose the maximum number of supervoxels to be 400, by visual inspection of the nature of the supervoxels that contain the tumour and the white matter hyperintensities for the two datasets. For example, between a setting where one supervoxel is part tumour and part background, versus another where one supervoxel fully represents tumour, we choose the latter case.

We then use either the supervoxels or simple cuboids (for the baseline methods) as areas to be masked, and the question arises of how many and how large areas to choose as masks to construct synthetic images for $D_{inp}$. Too small a volume, and it might be trivial for a network to inpaint it; too large, and it might not be a feasible task. For our experiments, we choose masks whose volume is at least 1500 voxels. For constructing cuboids, we randomly generate cuboids which are at least 12 units in each dimension (as $12^3$ is more than 1500, but $11^3$ is not). Finally, we ensure that the size of $D_{inp}$ is roughly 10 times that of the $D_{train}$, by choosing masks which fit the volume criteria as they are generated, and producing at most 10 synthetic images on-the-fly for a single real input image.

\noindent\textbf{Network Parameters}: The input size to the 3D U-net is 160 $\times$ 216 $\times$ 32, such that each input image is centre-cropped to 160 $\times$ 216 ($X-Y$ axes) to tightly fit the brain region in the scan, while we use the overlapping tile strategy in the Z-axis as inspired by the original U-net. Each of the 3 resolution levels consists of two 3 $\times$ 3 $\times$ 3 convolution layers using zero-padding and \emph{ReLU} activation, except for the last layer which is linear in the inpainter U-net and \emph{sigmoid} in the fine-tuned U-net. The number of feature channels are 16, 32 and 64 at the varying resolution levels. The feature maps in the upsampling path are concatenated with earlier ones through skip-connections.

\noindent\textbf{Optimization Parameters}: The inpainter U-net is optimized on MSE while fine-tuning is performed using a Dice objective, both using \emph{Adam}. The learning rate is 0.0001 and 0.001 for BraTS and WMH datasets, respectively. We used a batch-size of 4, as permitted by our GPU memory. For pre-training, we use 100 epochs while for fine-tuning we employ another 150, both without the possibility of early stopping, saving the best performing model based on the validation loss at every epoch.

To foster open-science, all of the code will be released\footnote{url-masked-for-blind-review}.

\section{Results and Discussion}
\label{experiments:results}

To study the effect reduced dataset sizes on the proposed approach, experiments were performed on the full training dataset as well as smaller fractions of it. For BraTS, we perform experiments on 25\%, 50\% and 100\% of the training data, while for WMH, which is much smaller in size, we only perform an extra set of experiments with 50\% of the data. To keep the comparisons fair, we use the same subset of the training data in the pre-training procedure as well.

The segmentation results are shown in Table \ref{t1}. It can be observed that the proposed method (\emph{roi-supervoxel}) outperforms the basic U-net (\emph{vanilla-unet}) by a large margin, and traditional inpainting based pre-training (\emph{noroi-grid}) by a small, but significant, margin.

The deductions from the empirical results can be summarized as follows:

\textbf{Restarts improve U-net performance:} It can be observed that for both datasets, the performance of the \emph{restart-unet} is better than that of the \emph{vanilla-unet}. This is in line with observations in literature \cite{sgdr}, where warm restarts have aided networks to find a more stable local minimum. Based on this observation, we argue that for any proposed method involving pre-training models, the results should always be compared to such a \emph{restarted} model.

\textbf{Adding ROI information to the inpainting proxy task is helpful:} For both the datasets, the performance of the \emph{roi-supervoxel} method exceeds that of all other baselines. Importantly, it exceeds the performance of the \emph{restart-unet} and the \emph{noroi-grid}, which is the traditional inpainting procedure, by 3.2\% and 5\% (relative) respectively for BraTS, and 4.9\% and 2.9\% for WMH, when all of the data is used. Also important to note is that the performance of methods which use the region-of-interest information to generate the masked areas is always better than those which do not.

\begin{table}[!t]
\footnotesize
\centering
\caption{\textbf{Dice-scores on BraTS 2018 and WMH 2017.} The results represent the mean and standard deviation (in brackets) of the Dice coefficient averaged over the three folds. The top results which are not significantly different from each other are depicted in \textbf{bold}.}
\begin{tabular}{|c|clclc|ccc|}
\hline
\multicolumn{9}{|c|}{{\ul \textbf{Fraction Training Data}}}                                                                                                    \\ \hline
\multicolumn{1}{|l|}{}    & \multicolumn{5}{c|}{{\ul \textbf{BraTS}}}                                & \multicolumn{3}{c|}{{\ul \textbf{WMH}}}                 \\ \hline
{\ul \textbf{Method}}     & \textbf{.25}         &  & \textbf{.50}         &  & \textbf{1.0}         & \textbf{.50}         & \textbf{} & \textbf{1.0}         \\ \hline
\textbf{vanilla-unet}     & 0.257 (.05)          &  & 0.585 (.03)          &  & 0.784 (.02)          & 0.576 (.05)          &           & 0.745 (0.02)         \\ \hline
\textbf{restart-unet}     & 0.302 (.05)          &  & 0.607 (.03)          &  & 0.793 (.02)          & 0.610 (.05)          &           & 0.776 (.03)          \\ \hline
\textbf{noroi-grid}       & 0.311 (.06)          &  & 0.611 (.04)          &  & 0.780 (.03)          & 0.632 (.04)          &           & 0.791 (.03)          \\ \hline
\textbf{roi-grid}         & \textit{0.354 (.06)} &  & 0.620 (.04)          &  & \textit{0.795 (.03)} & 0.653 (.04)          &           & \textbf{0.812 (.03)} \\ \hline
\textbf{noroi-supervoxel} & 0.340 (.06)          &  & \textit{0.621 (.04)} &  & 0.791 (.02)          & 0.650 (.04)          &           & 0.797 (.03)          \\ \hline
\textbf{roi-supervoxel}   & \textbf{0.363 (.06)} &  & \textbf{0.646 (.04)} &  & \textbf{0.814 (.03)} & \textbf{0.671 (.04)} &           & \textbf{0.814 (.03)} \\ \hline
\end{tabular}
\label{t1}
\end{table}

\textbf{Inpainting is more beneficial when the size of the training set is smaller:} The difference in performance between the inpainting-assisted methods and \emph{vanilla-unet} is larger when the size of the training dataset is smaller. For example, for BraTS, the difference between \emph{vanilla-unet} and \emph{roi-supervoxel} (our proposed approach) is as large as 41.2\% (relative) when the size of the training dataset is 25\% of the total. This trend is also observed between the methods with and without ROI information.

\textbf{Supervoxels help more when areas to be segmented are larger rather than finer:} ROI-guided inpainting can be postulated to have a better chance of affecting the downstream performance when the ROI itself is larger. Taking into account that tumours in BraTS are, on-average, larger than the hyperintensities to be segmented in the WMH dataset, it can be observed that the performance difference between the inpainting methods using supervoxels (\emph{roi-supervoxel} and \emph{noroi-supervoxel}) versus the ones which do not (\emph{roi-grid} and \emph{noroi-grid}) is smaller in the case of WMH than for BraTS. For example, when using all of the training data, the difference in performance between \emph{roi-supervoxel} and \emph{roi-grid} is 3\% (relative) for BraTS but only 0.25\% for WMH. This could likely be alleviated by problem specific selection of parameters for SLIC, which we did not explore. This would ensure that the supervoxels are not too large as compared to the ROI, in which case the effect of ROI would not be significant.

These results show that our approach is promising. An important point to note is that a similar approach may be valuable in other forms of local self-supervision techniques like jigsaw puzzle solving \cite{DBLP:conf/miccai/ZhuangLHMYZ19}, where the shuffling could be guided by the ROI and the tiles could be picked by ensuring homogeneity constraints.

Although efficient, this method does have some limitations: firstly, its efficiency depends on the parameters of the supervoxelization process and a poor choice of parameters could lead to limited performance gain; secondly, although sizeable synthetic datasets can be created in this process, the reliance on ROI means that we still need segmentation annotations. Perhaps one way of solving the second problem would be using co-training \cite{10.1145/279943.279962} to label all of the data and then employ our method using the entire corpus.

\section{Conclusion}
\label{conclusion}

In summary, this work explores the use of supervoxels and foreground segmentation labels, termed the \emph{region-of-interest (ROI)}, to guide the proxy task of inpainting for self-supervision. Together, these two simple changes have been found to add a significant boost in the performance of a convolutional neural network for segmentation  (as much as a relative gain of 5\% on the BraTS 2018 dataset), in comparison to traditional methods of inpainting-based self-supervision.

\bibliographystyle{splncs04}
\bibliography{mybibliography}

\end{document}